%% file: main.tex
\documentclass[10pt,twocolumn,letterpaper]{article}

\usepackage{wacv}
\usepackage{times}

\usepackage[noend]{algpseudocode}
\usepackage{microtype}
\usepackage{url}
\usepackage{subfiles}
\usepackage{graphicx}
\usepackage{epsfig}
\usepackage{multirow}
\usepackage{hhline}
\usepackage{amsmath}
\usepackage{amssymb}
\usepackage{enumitem}
\usepackage[toc,page]{appendix}
\usepackage{adjustbox}
\usepackage{xspace}
\usepackage{xcolor}
\usepackage{cite}
\usepackage{algorithm}
\usepackage{mathtools}
\usepackage[accsupp]{axessibility}

\usepackage[pagebackref=true,breaklinks=true,colorlinks,bookmarks=false]{hyperref}

\DeclareMathOperator*{\argmax}{arg\,max}
\DeclareMathOperator*{\argmin}{arg\,min}

\def\BLUERANK#1{{}{\color{blue}{#1}}}

\def\Vec#1{{\boldsymbol{#1}}}
\def\Mat#1{{\boldsymbol{#1}}}
\newtheorem{remark}{Remark}

\def\wacvPaperID{***} 

\wacvfinalcopy 

\ifwacvfinal
\fi

\ifwacvfinal
\usepackage[breaklinks=true,bookmarks=false]{hyperref}
\else
\usepackage[pagebackref=true,breaklinks=true,colorlinks,bookmarks=false]{hyperref}
\fi

\ifwacvfinal
\fi

\begin{document}

\title{Meta-Learning for Multi-Label Few-Shot Classification}

\author{%
\vspace{0.3cm}
  Christian Simon$^{\dagger, \S}$, \quad Piotr Koniusz$^{\S,\dagger}$, \quad Mehrtash Harandi$^{\clubsuit, \S}$\\\vspace{0.3cm}
  $^{\dagger}$The Australian National University \quad $^{\clubsuit}$Monash University \quad
   $^\S$Data61-CSIRO\\
  firstname.lastname\texttt{@\{anu.edu.au,monash.edu,data61.csiro.au\}} \\
}

\maketitle

\begin{abstract}
 Even with the luxury of having abundant data, multi-label classification  is widely known to be a challenging task to address. This work targets the problem of multi-label meta-learning, where a model learns  to predict multiple labels within a query (\eg, an image) by just observing a few supporting examples. In doing so, we first propose a benchmark for 
   Few-Shot Learning (FSL) with multiple labels per sample. Next, we discuss and extend several solutions specifically designed to address the conventional and single-label FSL, to work in the multi-label regime.
    Lastly, we introduce a neural  module to estimate the label count of a given sample by exploiting the relational inference. We will show empirically the benefit of the label count module, the label propagation algorithm, and the extensions of conventional FSL methods on three challenging datasets, namely MS-COCO, iMaterialist, and Open MIC.
   Overall, our thorough experiments suggest that the proposed label-propagation algorithm in conjunction with the neural label count module (NLC) shall be considered as the method of choice.
\end{abstract}

\input{sections/section1}
\input{sections/section2}

\input{sections/section3}
\input{sections/section4}

\input{sections/section5}

\input{sections/section6}

{\small
\bibliographystyle{ieee_fullname}
\bibliography{egbib}
}

\end{document}

%% file: sections/section1.tex
\section{Introduction}
\label{sec:intro}

Humans are able to learn novel concepts from very few examples~\cite{Hume1996LearningSetsConcepts,Lake2015Human}. 
Our capability of learning from limited examples is attributed to exploiting \emph{background knowledge}~\cite{Hume1996LearningSetsConcepts}. Such knowledge is presented as a collection of relationships, meaning that we are able to learn when various concepts are related and presented together. For example, we can identify a new species of fish, if its relation to concepts such as the sea and other aquatic plants and animals is explained. If the new species merely appears in the jungle among terrestrial animals, humans may fail to identify it without knowing how it relates to 
other animals.

\begin{figure*}
\centering
        \includegraphics[width=.95\textwidth]{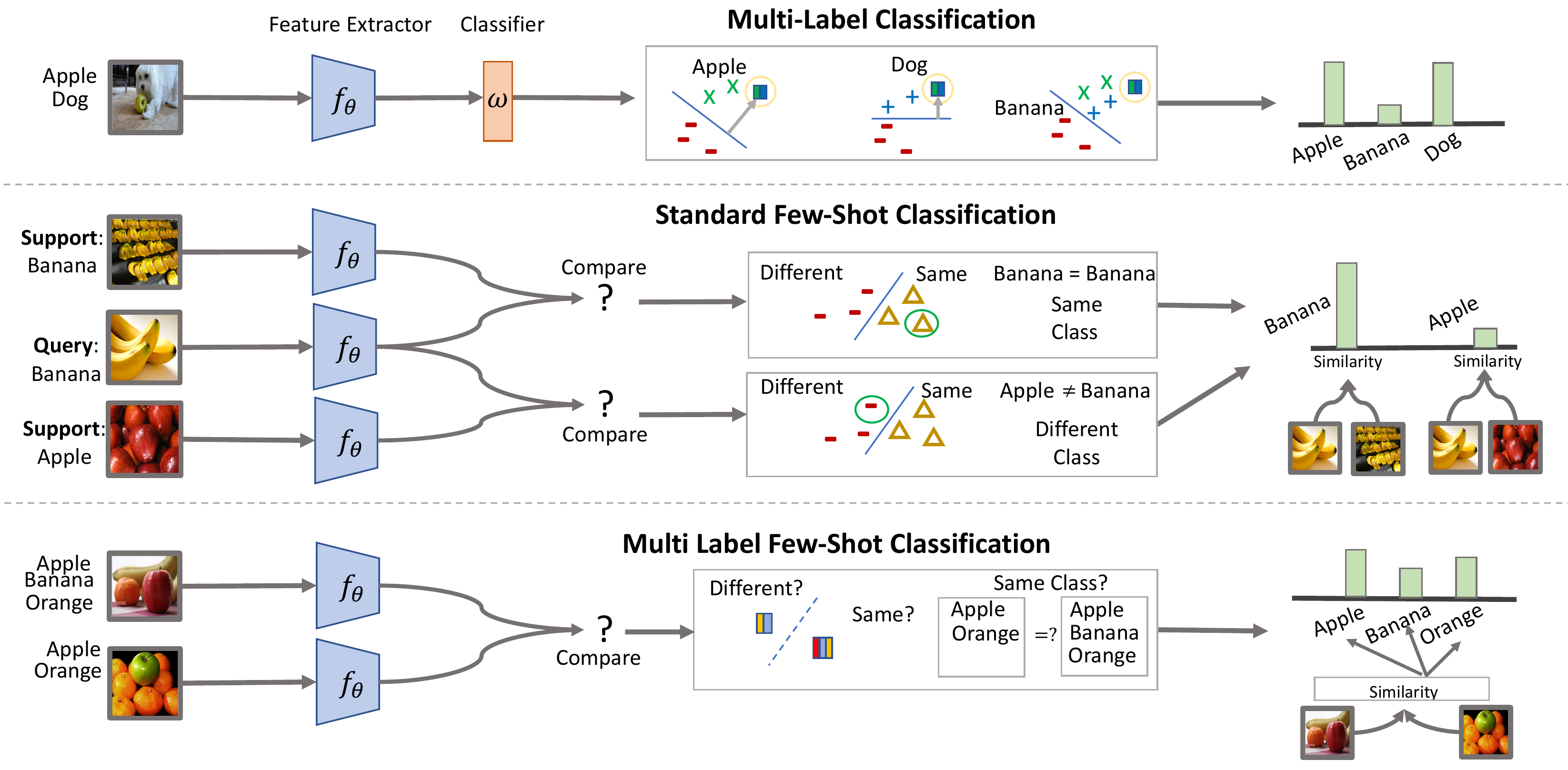}
\caption{Comparison between multi-label classification, single-label few-shot learning and multi-label few shot learning. Multi-label few-shot problem. \textbf{Top panel.} In classical multi-label classification, one  designs a fixed classifier from all seen classes. In a well-established practice, one breaks down the problem into identifying a set of binary classifiers where each classifier is responsible for identifying one specific class in a given input. 
\textbf{Middle panel.} A profound idea in addressing single-label FSL is to design a model to perform relational comparisons. 
Here, the model will receive pairs of images (the query and  an image from the support set) and predicts whether they are similar or not. We note that while in multi-label problems, the embedding of the query image is fixed, in relational inference, the embedding is dependent on the constructed pairs.
\textbf{Bottom panel.}
Extension of the relational inference to multi-label FSL regime is not trivial. 
Our work provides various solutions to address this challenging, yet extremely important and practical problem.
}
\label{fig:caseproblem} 
\end{figure*}

In machine learning, the so-called domain adaptation \cite{office_calt10,samita_domain,koniusz2017domain,zhang2018artwork,Koniusz2018Museum,action_da}, zero-shot learning \cite{romera2015embarrassingly,zhang2018zero,zhang2018model}, unsupervised/contrastive learning \cite{ssgc,zhang2021iept,refine,coles}, and few-shot learning~\cite{Santoro2016MANN,Vinyals2016MatchingNetworks,sosn,zhang2019few,9376901,christian_subs,Zhang_2020_ACCV,arl,zhang2020few,Christian2020ModGrad,9115215,zhang2019canet,zhu2021fewshot,Zhang_2020_CVPR,kon_tpami2020a}  help learning novel concepts from limited data.
New samples can also be  hallucinated with GANs \cite{GAN,NEURIPS2018_4e4e53aa,ShiriYPHK19,fatima_ijcv}. 
Despite the success, the conventional form of episodic learning comes with a  limiting assumption of 
being a single-label problem.
In other words, though multi-class, each data sample can only belong to one of possible classes. In the case of images, this means that each image should just encapsulate one visual concept (\ie, object). This  clearly limits the application of the developed techniques in many places (\eg, 
fashion recognition~\cite{Inoue2017Fashion}, multimedia content analysis~\cite{Qi2007MLVideo,Pachet2009Improving}, bioinformatics~\cite{Barutcuoglu2006HierMLPred,Cesa2012Synergy}, and drug discovery~\cite{Heider2013MultiDrug} 
to name a few). We  bridges this gap by introducing techniques to address Multi-Label Few Shot Learning (ML-FSL).

To get a feeling about the difficulty of ML-FSL, we recall that a profound idea in single-label FSL  is to make use of  relational comparisons. In essence, one learns to measure the similarity between pairs of samples, where a pair constitutes of the query and an example from the supporting set. While  comparing samples in the case of single-labels is well-behaved, extension to the multi-label regime is not an easy ride (see Fig.~\ref{fig:caseproblem} for a conceptual diagram).
This is because, by merely inspecting the similarity between pair of samples, one cannot deduce the class labels of the query.
In this work, we extend and introduce novel methods 
to tackle the problem of ML-FSL. In particular, we investigate: 
\begin{enumerate}
    \item Multi-Label Prototypical Networks. Inspired by the work of Snell \etal~\cite{Snell2017PrototypicalNetwork}, we make use of the notion of class prototypes to tackle the problem of ML-FSL. 
    \item Multi-Label Relation Networks. Building upon the work of Sung \etal~\cite{Sung2018Relnet}, we introduce multi-label relational networks with binary relevance.
    \item Label Propagation Networks. We propose a label propagation algorithm to address the problem of ML-FSL. Label propagation models the data in the form of graphs. This model learns the correlations among samples by weighting mechanism according to their similarities. 
\end{enumerate}

To the best of our knowledge, only the work of Alfassy \etal~\cite{Alfassy2019Laso} tackles the  problem of ML-FSL. Our work goes beyond this work in various ways. 
This includes, extending and introducing new algorithms for ML-FSL, developing a comprehensive evaluation framework with three challenging datasets, namely MS-COCO~\cite{Lin2014Coco}, iMaterialist~\cite{Guo2019Imaterialist}, and Open MIC~\cite{Koniusz2018Museum}
and proposing a neural model to perform label count, 
a module that empirically seems to be boosting the accuracy by a tangible margin.  

To summarize, our contributions in this paper are: 
\renewcommand{\labelenumi}{\roman{enumi}.}
\vspace{-2mm}
\hspace{-1cm}
\begin{enumerate}[leftmargin=1cm]
\item By departing from single-label FSL problems, we generalize and introduce various techniques to address the ML-FSL problem. 
\item A comprehensive and challenging evaluation framework for ML-FSL is developed.

\item We propose a neural label count module (NLC) to estimate the number of labels in an input and thoroughly asses its efficiency in conjunction with the proposed ML-FSL solutions.

\end{enumerate}

%% file: sections/section2.tex
\section{Problem Definition}
\label{sec:problemset}

Our goal is to construct meta-learning tasks such that the models can gain experience or learn from  similar tasks. Inspired by~\cite{Vinyals2016MatchingNetworks}, we propose the concept of multi-label few-shot classification via learning from \textit{episodes}. Moreover, the finite set estimation (label count) is also considered in this setting to complement the mAP measurement. Below, we explain our setting, notations, training, and testing strategies for ML-FSL. 

\paragraph{\textbf{Notations.}} Throughout this paper, we use bold lower-case letters (\eg, $\Vec{x}$) to denote column vectors and bold upper-case letters (\eg, $\Mat{X}$) to denote matrices.  A network is denoted by $f_\Theta:\mathbb{R}^m \xrightarrow{} \mathbb{R}^n$ which maps an input into some feature space. The set of labels per \textit{episode} is represented by $\mathcal{C}$.  Let $\mathcal{X} = \{ (\Vec{x}_1, \Vec{y}_1), \dots, (\Vec{x}_{N_x}, \Vec{y}_{N_x})) \}$  with $\Vec{x}_i \in \mathbb{R}^n; \Vec{y}_i \in \mathbb{R}^{|\mathcal{C}|}$ and  $\mathcal{Q} = \{\Vec{q}_1, \dots, \Vec{q}_{N_q}\}$ with $\Vec{q}_i \in \mathbb{R}^n$ denote the support set and query set, respectively. Element-wise index in a vector is denoted as $\Vec{x}_{(j)}$.

\begin{figure}
\centering
        \includegraphics[width=.44\textwidth]{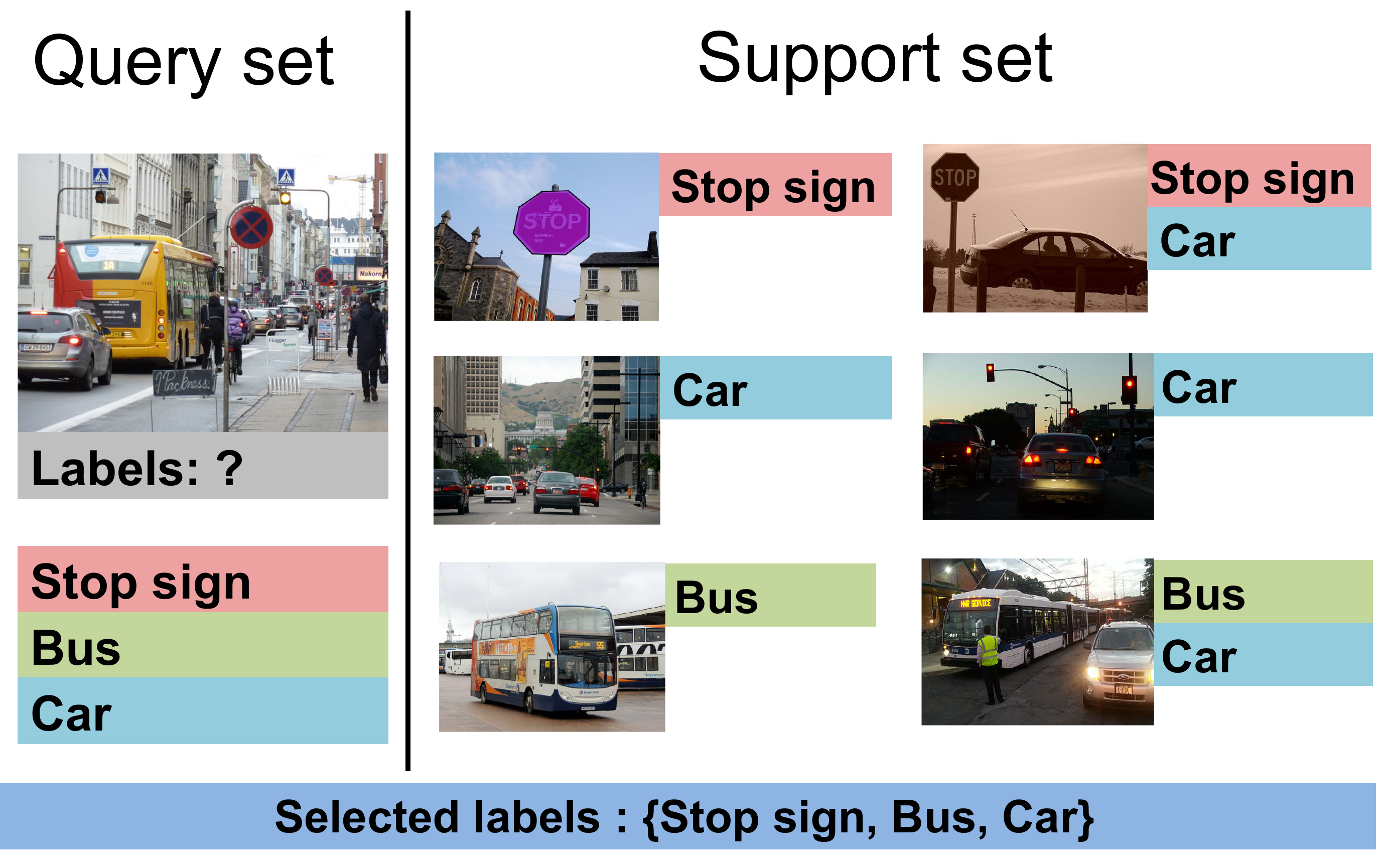}
\caption{An episode consists of a query set and a support set. The support set contains examples from selected classes of a given task. The query set covers the same set of labels presented in the support set.
}\vspace{-0.2cm}
\label{fig:setting} 
\end{figure}

\noindent \textbf{\textit{Episode}}. To form an episode (see Fig.~\ref{fig:setting}), we sample from a task distribution $T$ over possible label sets.
There are two sets shaping an \textit{episode}: a support set $\mathcal{X}$ and a query set $\mathcal{Q}$.  To address the multi-label classification, $\Vec{y}_i \subseteq \mathcal{C}$ denotes a set of multiple labels assigned to $\Vec{x}_i$. 
Note that, an episode composition including selected classes and examples may differ from one episode at a given timestep to another. This \textit{episode} structure exploits meta-knowledge such that a model that has the meta-learning capability will learn to match representations regardless of the actual semantic meaning of the labels, thus learning a relation between data points. As a result, the models can generalize and predict the appropriate labels given only a few data points during training. The challenges  such a technique has to address are twofold: (i) the number of data points is low and (ii) the size of the label set varies from image to image thus making the prediction task harder. 

\noindent \textbf{$N$-way $K$-shot}. In single-label few-shot learning, the term  $N$-way $K$-shot is used to describe the number of classes (way) in an episode and the number of examples (shot) in each class. 
For example, suppose there are three sampled classes ($N=3$), then we obtain five examples ($K=5$) per class to compose an \textit{episode}, so the support set contains 15 images in total.
Following this notation, we also sample $N$-way and $K$-shot from each label to form an \textit{episode} in ML-FSL.  However, the support set composition for ML-FSL differs from the single-label setting in the sense that the distribution over samples per class in an episode is not uniform. Because each data point $\Vec{x}_i$ in a support set can contain more than one label from the set of labels $\mathcal{C}$, thus, it is not guaranteed that there are exactly $K\times N$ samples constituting the support set of an  episode.

\noindent \textbf{Training Stage}. In majority of cases, to perform multi-label classification, pre-trained CNNs are used ~\cite{Wang2016CnnRnn,Wang2017MultiAttenReg,Zhu2017SpatialReg}. However, in this paper, we want to focus on the learning techniques for which the network must exhibit the capacity to improve its performance by learning from previous tasks. As a result, we train the networks from scratch by random initialization. The models are updated based on the training objective as follows: 
\begin{equation}
\resizebox{0.9\hsize}{!}{
$\Theta^\ast = \argmax_{\Theta} \mathbb{E}_{\mathcal{C} \sim T}\Bigg{[} \mathbb{E}_{\mathcal{X} \sim \mathcal{C},\mathcal{Q} \sim \mathcal{C}}\Bigg{[} \sum_{\Vec{q} \in \mathcal{Q}} \text{log} P_{\Theta}(\Vec{y} | \Vec{q},\mathcal{X}) \Bigg{]}\Bigg{]}$,}
\end{equation}
where $\Theta$ denote the model parameters that we want to learn. The model is trained over many episodes to minimize the prediction error over  the query set $Q$. 

\noindent \textbf{Testing Stage}. In the testing stage, the model performs  classification via assigning each data point to unseen labels from a distribution $T^\prime$. Following an \textit{episode} composition in the training stage, there are $N$ labels and at least $K$ examples per label in the support set. Thus, one can think of such classification strategy as transfer learning from previously learned tasks to the new set of tasks.  

%% file: sections/section3.tex
\section{Meta-Learning for Multi-Label Few-Shot Problems}
\label{sec:method}

\begin{figure*}
\centering
        \includegraphics[width=.75\textwidth]{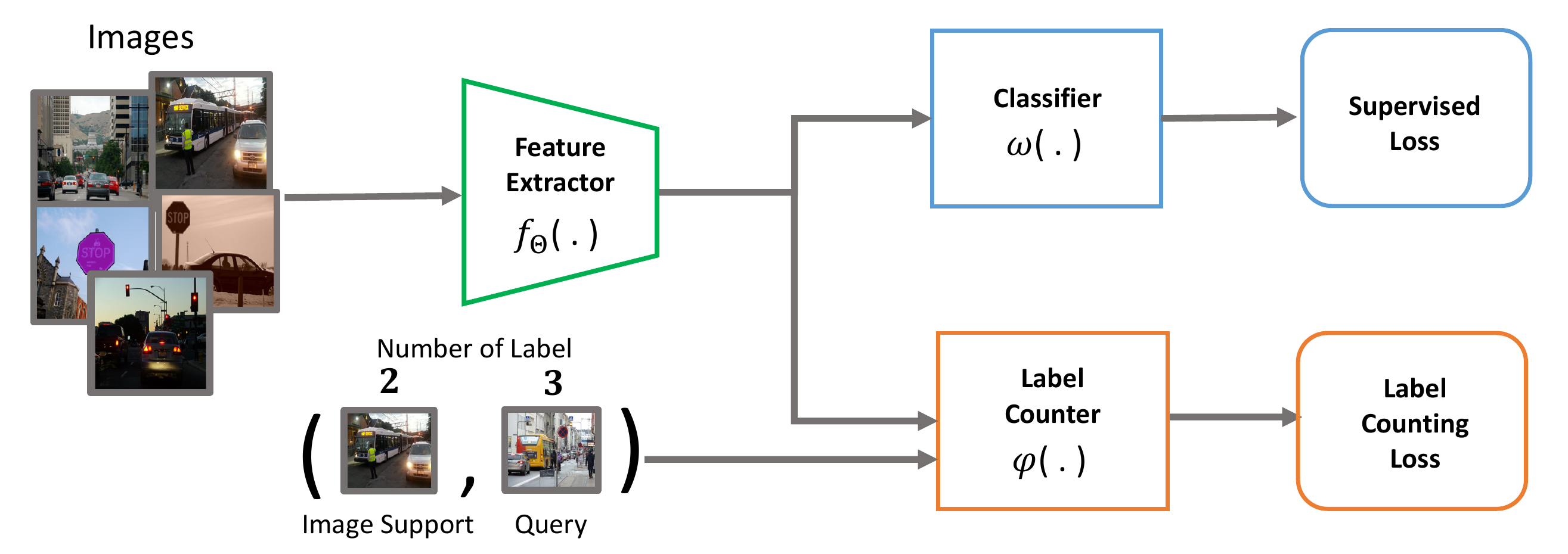}
\caption{The general architecture to classify and predict the number of labels in training.}
\label{fig:architecture} 
\end{figure*}

In this section, we first extend three successful FSL schemes to their ML-FSL counterparts (see Fig.~\ref{fig:architecture} for a conceptual diagram). This is followed by describing the proposed NLC module. 

\subsection{From Single-Label to Multi-Label}
\label{subsec:singlelabelmethods}
Below, we extend and adapt prototypical networks~\cite{Snell2017PrototypicalNetwork}, relation networks~\cite{Sung2018Relnet}, and label propagation~\cite{Liu2018TPN} to their multi-label versions.

\noindent\textbf{Prototypical Networks.}
Prototypical networks  
learn a mapping $f_\Theta:\mathbb{R}^m \xrightarrow{}\mathbb{R}^n$
from an input space to an embedding space. The main assumption is that the embeddings should sit near their class prototypes. Let us denote the prototypes (for $C$ classes in an episode) by $\Mat{P} = \{\Vec{p}_1, \dots, \Vec{p}_C\}; \Vec{p}_k \in \mathbb{R}^n$ and $\Mat{X}_i$ be a class-specific set. A prototype is calculated as the mean of the feature vectors
that share the same class label:
\begin{equation}
\Vec{p}_k = \frac{1}{|C_k|}\sum_{\Vec{x}_j \in \Vec{X}_k} f_\Theta(\Vec{x}_j).
\end{equation}
In prototypical networks, a query feature is classified by assigning it to the nearest prototype according to the Euclidean distance: $\Vec{z}_{(j)} = \|\Vec{p}_j - f_\Theta(\Vec{q})\|^2$.

We adapt prototypical networks to work in the multi-label setting. For a given class, all samples annotated with that class are grouped together to form a prototype. Note that every sample with more than one label will contribute in formation of several prototypes. To perform a multi-label classification, we conduct training with a softmax function~\cite{Yang2016Exploit,Wang2017MultiAttenReg} in the final layer of network. The objective function is formulated as follows:
\begin{equation}
\mathcal{L}_{PN} = \sum_{j=1}^{|\mathcal{C}|} \Big{(}\frac{\Vec{y}_{(j)}}{\|\Vec{y}\|_1} - \frac{\text{exp}( \Vec{-z}_{(j)} )}{\sum_{j^\prime=1}^{\mathcal{|C|}}\text{exp}(\Vec{-z}_{(j^\prime)})}\Big{)}^2.
\end{equation}

\noindent\textbf{Relation Networks.} 
Few-shot recognition can also be performed via the use of the so-called \textit{relation module} from relation networks (RN). This approach can be viewed as learning deep
non-linear metric. By and large, RN consists of an embedding module and a \textit{relation module}. Specifically, an embedding module is a non-linear mapping from the input space to a feature space $f_\Theta:\mathbb{R}^m \xrightarrow{}\mathbb{R}^n$ and the \textit{relation module} ($g_\Phi$) learns the similarity between the query  ($\Vec{q}$) and sample ($\Vec{x}_j$) in the support set. Let us denote $\Mat{X}_i$ be a class-specific set. Given a problem with $\mathcal{C}$ classes, the relation scores can be calculated according to:
\begin{equation}
    \Vec{r}_{(j)} = g_\Phi (f_\Theta(\Vec{q}), \frac{1}{|C_j|}\sum_{\Vec{x}_i \in \Mat{X}_j} f_\Theta(\Vec{x}_i)), \quad j = 1, \dots, \mathcal{C} .
\end{equation}
Architecture-wise, the \textit{relation module} is comprised of two convolutional blocks, two fully connected layers, and a sigmoid function ($\sigma$).
Training is performed by calculating mean squared error between the score and the query label w.r.t. the model parameters and embeddings:
\begin{equation}
    (\Theta, \Phi) = \argmin_{\Theta, \Phi} \sum_{j=1}^{|\mathcal{C}|} (\Vec{r}_{(j)} - \Vec{y}_{(j)})^2.
\end{equation}
Thus, RN passes the features from the support set and a query to the binary classifier and  the label  with highest score is chosen as the predicted class.

Adapting RN from single-label to multi-label setting is straightforward. The \textit{relation module} acts as a non-linear metric, thus, we can directly use a log-loss 
w.r.t.  $\Vec{r}_{(j)}$ and labels $y_j$: 

\begin{equation}
     \mathcal{L}_{RN} =\sum_{j=1}^{|C|} \Vec{y}_{(j)} \text{log}(\Vec{r}_{(j)}) + (1 - \Vec{y}_{(j)})\text{log}(1 - \Vec{r}_{(j)}).
\end{equation}

\noindent\textbf{Label Propagation.}
Another approach is to model few-shot learning with a graph and make use of label propagation methods to classify a query sample. To predict the concept scores, we make use of  the \emph{\textbf{smoothness}} property, meaning that similar instances should have similar concept scores. More specifically, if a query sample $\Vec{q}_j$ is similar to a support instance $\Vec{x}_i$, then $\Vec{\psi}_j \approx \nu(\Vec{y}_i)$ where $\nu:\mathbb{R}^C \to \mathbb{R}^C$ is the $\ell_1$ normalization operator defined as $\nu(\Vec{y}) = \Vec{y}/ \|\Vec{y}\|_1$. Let $\mathbb{R}^{n \times (N_s + N_q)} \ni \Mat{X} = [\mathcal{X},\mathcal{Q}] = [\Vec{x}_1,\cdots,\Vec{x}_{N_s},\Vec{q}_1,\cdots,\Vec{q}_{N_q}]$. We denote the columns of $\Mat{X}$ with $\Vec{x}_i; 1 \leq i \leq N_s + N_q$. Similarly, define 
$\mathbb{R}^{C \times (N_s + N_q)} \ni \Mat{\Phi} =  [\nu(\Vec{y}_1),\nu(\Vec{y}_2),\cdots,\nu(\Vec{y}_{N_s}),\Vec{\psi}_1,\Vec{\psi}_2,\cdots,\Vec{\psi}_{N_q}]$. $\Mat{\Phi}_{\mathcal{X}}$ has a similar structure with $\Vec{\psi}_i = 0, i = N_s + 1,\cdots, N_s + N_q$.

To achieve our goal, we start by building a neighborhood graph ($W$) over the support and the query sets using $\Mat{X}$ according to 
\begin{equation}
    w_{i,j} = \begin{cases}
    \begin{aligned}[b]
    \exp\big(-\sigma \|f_\Theta(\Vec{x}_i) - f_\Theta(\Vec{x}_j)\|^2\big)  \end{aligned},  & j \in \mathcal{N}_i \\
    \\ 0, & \mbox{otherwise}.
    \end{cases}
\label{eq:graphweight}
\end{equation}
Here, $\mathcal{N}_i$ is the set of neighbors of node $i$ that share the same class label to $\Vec{x}_i$.
Then, a normalized graph is formed with $\Mat{S} = \Mat{D}^{-\frac{1}{2}}\Mat{W}\Mat{D}^{-\frac{1}{2}}$ where $\Mat{D}$ is a diagonal matrix with $\Mat{D}_{ii} = \sum_j \Mat{W}_{ij}$. Then final prediction scores has a closed form 
(see~\cite{Zhou2004GraphClosed} for example) and can be calculated:
\begin{equation}
    \Mat{F}^{*} = \Mat{\Phi}_{\mathcal{X}}(\mathbf{I} - \alpha \Mat{S})^{-1},
\end{equation}
where $\alpha$ is usually set to 0.99 in practice. The final classification loss for multi-label setting is defined as:
\begin{equation}
     \mathcal{L}_{LP} = \big{\|} \Mat{\Phi} - \Mat{F}^{*} \big{\|}^2.
     \label{eq:finalloss_supp}
\end{equation}

\subsection{Neural Label Count}
We begin with the role of self-supervision learning for the multi-label problem. The characteristic of self-supervision training is to use the characteristic of the data. For instance, by rotating an image, one can self-supervise a machine by predicting the amount of rotation. Here, we make use of the arithmetic operation for prediction. 
The functionality of the NLC module is self-explanatory. The module predicts the number of classes (\eg, objects in an image) presented in a given input.
Our design benefits from a relation module that takes into account a support sample, the query, and the global information of a task, represented in the support set . The function to predict the number of labels is denoted by: 
\begin{equation}
  M^{[\Vec{x_i,q}]} = \varphi(f_\Theta(\Vec{x_i}),f_\Theta(\Vec{q}),\Vec{z}),
\end{equation}
where the multi-label counter function $\varphi:
\mathbb{R}^{d}\times\mathbb{R}^{d}\times\mathbb{R}^{d} \to \mathbb{R}^{2|C|}$ learns the relation between two samples (\ie, $f_\Theta(\Vec{x_i})$ and $f_\Theta(\Vec{q})$). 
Here, $\Vec{z}$ is a vector carrying the context of the whole support set to the NLC module. In our implementation, we realize this as $\Vec{z} = \frac{1}{NK} \sum_{\Vec{x}_i \in \mathcal{X}}{f_\Theta(\Vec{x_i})}$.

In particular and as becomes clear shortly, the NLC predicts the number of objects presented collectively in 
$\Vec{x_i}$ and $\Vec{q}$. In doing so, the NLC module 
uses a softmax with $2|C|$ outputs as the maximum number of 
classes presented collectively in $\Vec{x_i}$ and $\Vec{q}$
cannot exceed $2|C|$.

Our training OBJECTIVE is to minimize the following loss function:
\begin{equation}
    \mathcal{L}_{E} = \mathcal{L}_{su} + \lambda\mathcal{L}_{co},
\end{equation}
where $\mathcal{L}_{su}$ depends on the selection of the classifier and $\mathcal{L}_{co}$ is formulated as:
\begin{equation}
    \mathcal{L}_{co} = -\sum_j\sum_i\text{log}\Bigg{(}\frac{\text{exp}(  M^{[\Vec{x_i,q}]}_{(j)})}{\sum_{j^\prime}\text{exp}(M^{[\Vec{x_i,q}]}_{(j^\prime)})}  \Bigg{)}.
\end{equation}

\begin{figure}
\centering
        \includegraphics[width=.4\textwidth]{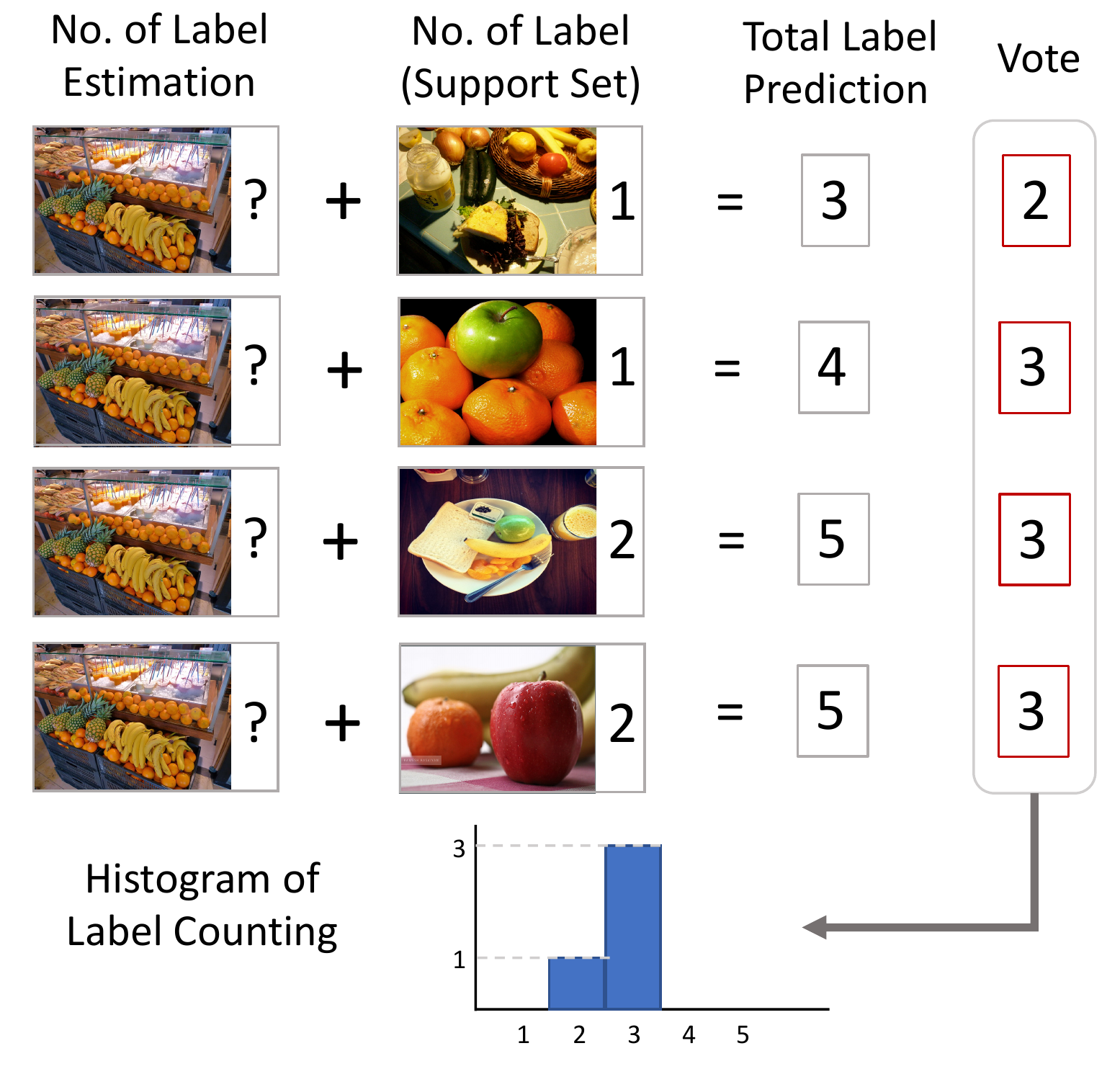}
\caption{Estimating the number of label by voting system of the support samples and queries. Here, an estimation (top) predicts the number of label as 2 but the rest predictions are 3.}
\label{fig:inferencevoting} 
\end{figure}

\subsection{Inference with Label Count Voting}
We employ a voting scheme on the output of the NLC module. 
To be more specific, we compare each support sample with the query and create the histogram of the label count estimation as shown in Fig.~\ref{fig:inferencevoting}.  The label counting based on histogram ($\Mat{H}$) is  defined as:
\begin{equation}
    \Mat{H} = \{ h(m) | 0 \leq m \leq 2|C|\},
\end{equation}
with
\begin{equation}
    h(m) = \sum_{\Vec{x_i} \in \mathcal{X}} \delta((M^{[\Vec{x_i,q}]} - B^{\Vec{x_i}}) - m),
\end{equation}
where ${B}^{\Vec{x}_i}$ is the label count of the support sample $\Vec{x}_i$ and $\delta(\cdot)$ returns 1 if its argument is equal 0 (or it returns 0 for the input not equal 0). Indeed, $h(m)$ shows how many times, the number of classes in the query is counted as $m$. 
A majority voting is then used to make the final call as:
\begin{equation}
    l = \argmax_{i \in \{1, \cdots, |C|\}} \, h(i).
\end{equation}
This label count estimation is also included in our evaluation scheme.

\begin{remark}
We consider the episode style for training and testing  multi-label few-shot classification. In this setting, the labels per episode are randomly picked and every episode may contain different samples. Thus, there is a chance that some objects are labeled in an episode but they are not labeled in another episode. This is also known as missing labels in multi-label classification. Based on this problem,
the relation module is designed to be an adaptive module that capture the information based on the context in an episode.  
\end{remark}

\begin{remark}
The NLC module has a different use compared to relation networks~\cite{Sung2018Relnet} that exploits the similarity between two features. The neural label count module conveys the operator relationship (\ie, summation ) such that the result of this relation module is the number after an arithmetic operation.
\end{remark}

\begin{remark}
This inference method makes use of the ensemble strategy to estimate the label count. The ensemble of label count estimations implies that a level of robustness might be expected. That is,  even if one of the classifiers is wrong in its prediction, the other predictions may correct it through consensus. The final prediction is made based on the most frequent label count predictions. In the case of a tie,  then the original predictions (floating point) are used to check the number with the highest probability.

\end{remark}

%% file: sections/section4.tex
\section{Related Work}
\label{sec:relatedwork}

In the multi-label setting, the main challenge is to classify an example belonging to multiple different classes simultaneously. The problem is challenging as an algorithm has to return a correct set of labels for previously unseen sample (the label count differs per sample).
The problem at hand is obviously more difficult as the solution needs to 
infer not only the most likely labels but also decide on their numbers. 

\noindent\textbf{Few-shot Learning.} 
Metric-learning approaches are proven beneficial to tackle classification tasks with low number of samples~\cite{Sung2018Relnet,Snell2017PrototypicalNetwork,Koch2015Siamese}. 
The embedding networks learn to compare contents of episodes in order to infer the underlying discriminative model for the given tasks. Initially, siamese~\cite{Koch2015Siamese} and matching networks~\cite{Vinyals2016MatchingNetworks} apply the idea of nearest neighbor inference for the task of few-shot classification. Prototypical networks~\cite{Snell2017PrototypicalNetwork} models the feature representations from samples within the same class as a single prototype. Furthermore, a relationship between class representations and queries can be also learnt through a neural network unit \eg, so-called Relation Networks~\cite{Sung2018Relnet}. Such a family of few-shot learning techniques learns an underlying metric, that is, the distance between any two feature vectors becomes small only if they belong to the same class. Another few-shot model using metric learning is through model adaptation as proposed in~\cite{Hong_2021_CVPR,Christian2020ModGrad}.  Another work uses graph as transductive label propagation networks in~\cite{Liu2018TPN}. These works are related to our extension from single-label few-shot learning to ML-FSL.

\noindent\textbf{Multi-label Classification.}
Focusing on deep learning solutions for multi-label classification, the majority of  studies use the so-called joint embedding models to solve the task in-hand~\cite{Frome2013Devise,Weston2011Wsabie}.
The multi-label classification problem can also be tackled in a sequential manner via a recurrent neural network (RNN)~\cite{Wang2016CnnRnn} which aggregates feature vectors produced by a CNN feature encoder. Recently, attention-based approaches for multi-label image recognition have been developed. An iterative attentional regions discovery is proposed in~\cite{Wang2017MultiAttenReg}. These attentional regions  correspond to semantic labels and thus they are embedded via LSTM. Moreover, a spatial regularization network (SRN)~\cite{Zhu2017SpatialReg} learns semantic and spatial label relations from attention maps. Thus, initial confidence scores can be adjusted via SRN to new regularized scores.
The correlation between two labels and an image feature obtained from CNN is expressed via correlation matrix. Then, the determinant of the correct subset is maximized to learn the deep neural network (DNN). Intuitively, one can think of this approach as maximizing a set of subspaces, each spanned by one ground-truth set of labels and corresponding features while minimizing the remaining sets of subspaces. Moreover, a previous work~\cite{Li2017Improving} uses pair wise ranking and threshold or label count estimation to improve the result of multi-label classification.

%% file: sections/section5.tex
\section{Experiments}
\label{sec:experiments}

\begin{figure}
\centering
        \includegraphics[width=.4\textwidth]{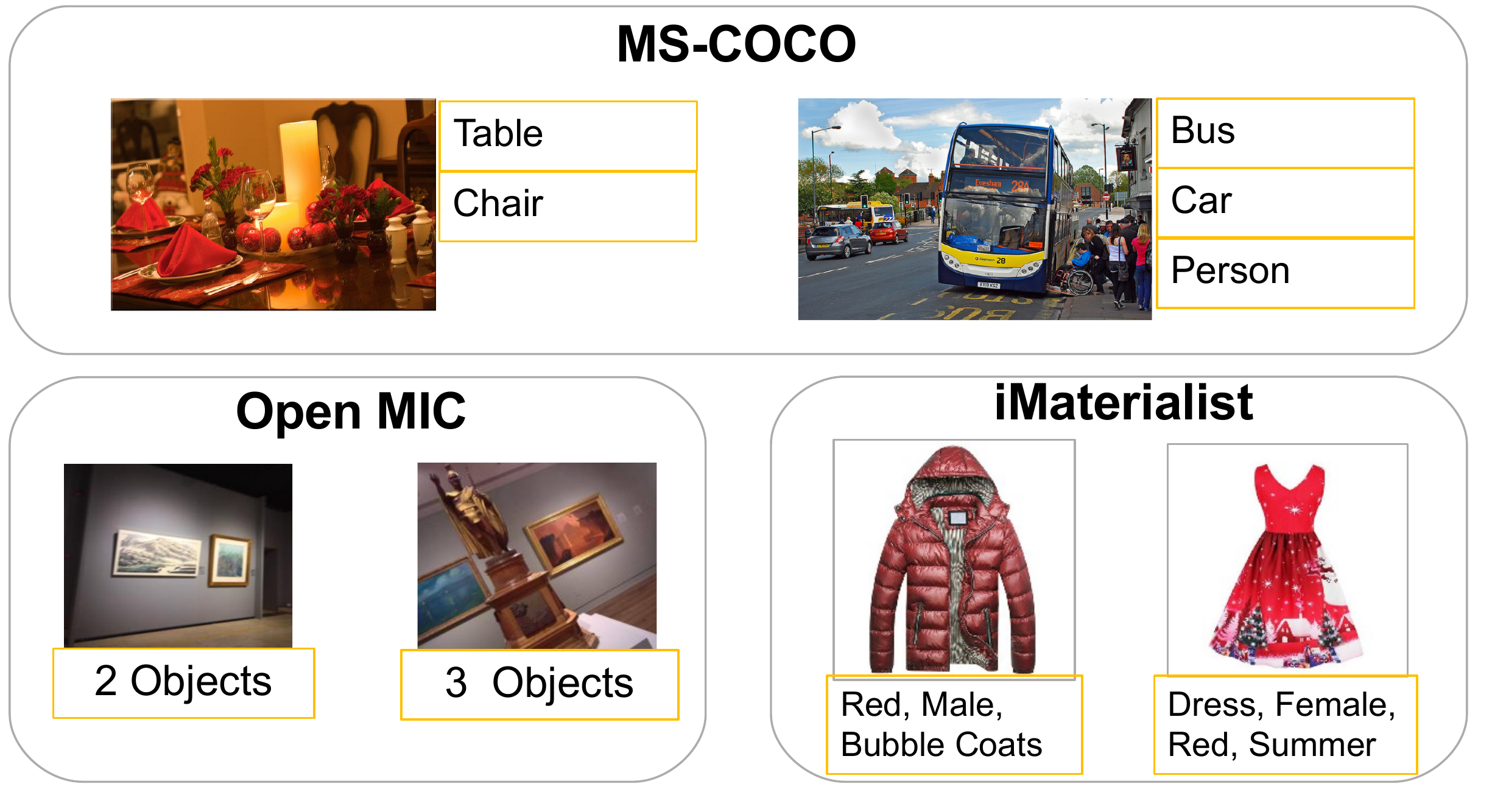}
\caption{Sample images from three datasets: MS-COCO, iMaterialist, and Open MIC.}
\label{fig:sampleimages} 
\end{figure}

\begin{table}[t]
    \centering
     \resizebox{.37\textwidth}{!}{
    \small\addtolength{\tabcolsep}{-2.5pt}

    \begin{tabular}{|c|c|c|}
    \hline
         \textbf{Model} & \textbf{1-shot} & \textbf{5-shot} \\ \hline
         LASO (intersection aug.) &40.5 &57.2\\
         LASO (union aug.) &45.3 &58.1\\
          \hline
          Proto Nets &{48.7} &{59.9}\\
          Relation Nets & 49.5 & 58.5\\
          LPN &{56.1} &{63.4}\\
          Proto Nets \texttt{+} NLC&\textbf{50.2} &\textbf{60.4}\\
          Relation Nets \texttt{+} NLC&\textbf{53.3} &\textbf{60.8}\\
          LPN \texttt{+} NLC&\textbf{56.8} &\textbf{64.8}\\
         \hline
    \end{tabular}
    }\vspace{0.2cm}
    \caption{A comparison in mAP(\%) to the existing benchmark~\cite{Alfassy2019Laso} on 16-way 1-shot and 5-shot. }
    \label{table:lasosetting}
\end{table}

\begin{table*}[t]
    \centering
    \resizebox{0.78\textwidth}{!}{
    \small\addtolength{\tabcolsep}{-.5pt}
    \begin{tabular}{|c|c c|c c|c c|c c|}
    \hline
         
         &\multicolumn{4}{c|}{\textbf{Base Classes}} &\multicolumn{4}{c|}{\textbf{Novel Classes}}  \\
        \cline{2-9}
       \textbf{Model} &\multicolumn{2}{c|}{\textbf{1-shot}}  &\multicolumn{2}{c|}{\textbf{5-shot}} &\multicolumn{2}{c|}{\textbf{1-shot}} &\multicolumn{2}{c|}{\textbf{5-shot}} \\
        \cline{2-9}
        &\textbf{mAP}  &\textbf{LC}  &\textbf{mAP} &\textbf{LC}  &\textbf{mAP}  &\textbf{LC}  &\textbf{mAP} &\textbf{LC} \\
        \hline
          Pre-trained \texttt{+} MLP  &$56.6$ &-  &$57.5$ &-  &$50.2$ &-  &$54.4$ &-  \\
        
        Proto Nets  &$61.0$ &-  &$69.7$ &-  &$56.7$ &-  &$66.7$ &- \\
        Relation Nets   &$64.4$ &- &$69.3$ &-  &$57.3$ &-   &$63.3$ &-  \\
        LPN  &${64.8}$ &-  &${71.8}$ &-  &${58.5}$ &-  &${68.3}$ &-  \\
        \hline
        
        Proto Nets \texttt{+} NLC &$62.8\uparrow$ &$\BLUERANK{40.2}$  &$\BLUERANK{72.3}\uparrow$ &$\BLUERANK{45.1}$  &$58.0\uparrow$ &$\mathbf{49.5}$  &$\BLUERANK{68.0}\uparrow$ &$54.1$  \\
        Relation Nets \texttt{+} NLC  &$\mathbf{66.2}\uparrow$ &$\mathbf{42.7}$  &$71.0\uparrow$ &$42.8$  &$\BLUERANK{59.0}\uparrow$ &$\BLUERANK{48.0}$   &$66.1\uparrow$ &$\mathbf{57.0}$  \\
        LPN \texttt{+} NLC&$\BLUERANK{66.0}\uparrow$ &$39.0$  &$\mathbf{74.5}\uparrow$ &$\mathbf{46.0}$  &$\mathbf{60.4}\uparrow$ &$47.4$  &$\mathbf{69.1}\uparrow$ &$\BLUERANK{54.6}$  \\

    \hline
    \end{tabular}

    }
    \vspace{0.2cm}
     \caption{The accuracy (\%) of baseline methods and the additional NLC on MS-COCO. The evaluation is based on mAP and LC for 10-way 1-shot and 5-shot. The best performance is in \textbf{bold} and the second best is in~\BLUERANK{blue} font.   } 
    \label{table:mscoco}
\end{table*}

\begin{table*}[!ht]
 \centering
    \resizebox{1.0\textwidth}{!}{
    \small\addtolength{\tabcolsep}{1.5pt}
\begin{tabular}{|c|c c|c c|c c|c c|}
\hline
     \multirow{3}{*}{\textbf{Model}} &\multicolumn{4}{c|}{\textbf{MS-COCO}} &\multicolumn{4}{c|}{\textbf{iMaterialist}}\\
     \cline{2-9}
     &\multicolumn{2}{c|}{\textbf{1-shot}} &\multicolumn{2}{c|}{\textbf{5-shot}}  &\multicolumn{2}{c|}{\textbf{1-shot}} &\multicolumn{2}{c|}{\textbf{5-shot}}\\ \cline{2-9}
     &\textbf{10-way} &\textbf{15-way}  &\textbf{10-way} &\textbf{15-way} &\textbf{10-way} &\textbf{15-way}  &\textbf{10-way} &\textbf{15-way}\\
      \hline
        Proto Nets \texttt{+} NLC &$30.4$ &$20.9$  &${37.1}$ &$24.8$  
        &$45.6$ &$\mathbf{43.6}$  &$47.4$ &$\mathbf{47.2}$ \\
        Relation Nets \texttt{+} NLC  &$26.2$ &$14.3$  &$29.2$ &$17.27$  
        &$\mathbf{46.5}$ &$42.7$  &$\mathbf{49.7}$ &$47.0$ \\
        LPN \texttt{+} NLC &$\mathbf{31.1}$ &$\mathbf{24.2}$  &$\mathbf{37.8}$ &$\mathbf{28.42}$
        &$45.4$ &$40.4$  &$48.9$ &$43.5$ \\
        \hline
\end{tabular}
}
\caption{The accuracy (\%) which quantifies multi-label hard predictions. The counted label is used in making  predictions based on thresholds. }
\label{table:harddecision_way}
\end{table*}

\subsection{Datasets}
We use three datasets detailed below to evaluate the capability of the methods to conduct ML-FSL. Note that, we have two disjoint sets for training and testing, so the images with annotated classes in the testing set do not appear in the training set. We propose the new splits for these three datasets. Fig.~\ref{fig:sampleimages} shows examples from all datasets.

\noindent\textbf{MS-COCO}~\cite{Lin2014Coco}. The MS-COCO dataset is built for object detection and recognition, thus, the data is suitable for multi-label recognition. This dataset comprises 80 classes and 122,000 images in total. We split the training, validation, and testing set into 50, 10, and 20 classes, respectively. We categorize 50 training classes as base classes and 20 testing classes as novel classes. All of the images within training are disjoint from validation and testing sets. Additionally, we remove images that have less than two labels yielding 74,655 images. We use all of the images from the validation set of MS-COCO to evaluate performance on base classes. 

In addition, we also evaluate on the MS-COCO split in~\cite{Alfassy2019Laso} to compare with the existing approach. This split has 64 and 16 classes for training and testing, respectively.

\begin{table}[t]
 \centering
    \resizebox{0.45\textwidth}{!}{
    \small\addtolength{\tabcolsep}{1.5pt}
\begin{tabular}{|c|c c |c c |}
\hline
     \multirow{3}{*}{\textbf{Model}} &\multicolumn{4}{c|}{\textbf{15-way}}\\
     \cline{2-5}
     &\multicolumn{2}{c|}{\textbf{1-shot}} &\multicolumn{2}{c|}{\textbf{5-shot}} \\ \cline{2-5}
     &\textbf{mAP} &\textbf{LC}  &\textbf{mAP} &\textbf{LC} \\
      \hline
        Proto Nets  &$48.4$ &-  &$59.8$ &-  \\
        Relation Nets   &$52.2$ &-  &$55.9$ &-  \\
        LPN  &${52.3}$ &-  &$\BLUERANK{61.1}$ &-   \\
        \hline
        Proto Nets \texttt{+} NLC &${49.4}\uparrow$ &$\BLUERANK{36.5}$  &${61.0}\uparrow$ &$\mathbf{45.0}$  \\
        Relation Nets \texttt{+} NLC  &$\mathbf{54.4}\uparrow$ &$\mathbf{41.1}$  &${54.4}\uparrow$ &${35.7}$  \\
        LPN \texttt{+} NLC &$\BLUERANK{53.6}\uparrow$ &${38.1}$  &$\mathbf{63.6}\uparrow$ &$\BLUERANK{43.6}$  \\
        \hline
\end{tabular}
}
\vspace{0.2cm}
\caption{The accuracy (\%) of baseline methods with and without the auxiliary NLC loss on the MS-COCO. The evaluation is based on mAP and LC for 15-way 1-shot and 15-way 5-shot protocols. The best performance is highlighted by the \textbf{bold} font and the second best by the \BLUERANK{blue} font. 
} 
\label{table:coco_15way}
\end{table}

\begin{table}[t]
 \centering
    \resizebox{0.45\textwidth}{!}{
    \small\addtolength{\tabcolsep}{1.5pt}
\begin{tabular}{|c|c c |c c |}
\hline
     \multirow{3}{*}{\textbf{Model}} &\multicolumn{4}{c|}{\textbf{15-way}}\\
     \cline{2-5}
     &\multicolumn{2}{c|}{\textbf{1-shot}} &\multicolumn{2}{c|}{\textbf{5-shot}} \\ \cline{2-5}
     &\textbf{mAP} &\textbf{LC}  &\textbf{mAP} &\textbf{LC} \\
      \hline
        Proto Nets  &$52.9$ &-  &$62.6$ &-  \\
        Relation Nets   &$\BLUERANK{58.4}$ &-  &$\BLUERANK{64.6}$ &-  \\
        LPN  &${56.4}$ &-  &${59.2}$ &-   \\
        \hline
        Proto Nets \texttt{+} NLC &${54.0}\uparrow$ &$\BLUERANK{28.0}$  &${64.1}\uparrow$ &$\mathbf{31.9}$  \\
        Relation Nets \texttt{+} NLC  &$\mathbf{59.9}\uparrow$ &${27.7}$  &$\textbf{65.3}\uparrow$ &$\BLUERANK{31.9}$  \\
        LPN \texttt{+} NLC &${57.4}\uparrow$ &$\mathbf{29.5}$  &${61.0}\uparrow$ &${29.5}$  \\
        \hline
\end{tabular}
}
\vspace{0.2cm}
\caption{The accuracy (\%) of baseline methods without and with the auxiliary NLC loss on iMaterialist. The evaluation is based on mAP and LC for the 15-way 1-shot and 15-way 5-shot protocols. The best performance is highlighted by the \textbf{bold} font and the second best by the \BLUERANK{blue} font. 
} 
\label{table:imat_15way}
    \vspace{-0.1cm}
\end{table}

\begin{table}[t]
 \centering
    \resizebox{0.45\textwidth}{!}{
    \small\addtolength{\tabcolsep}{-2.5pt}
\begin{tabular}{|c|c c |c c |}
\hline
     \multirow{2}{*}{\textbf{Model}} &\multicolumn{2}{c|}{\textbf{1-shot}} &\multicolumn{2}{c|}{\textbf{5-shot}} \\ \cline{2-5}
     &\textbf{mAP} &\textbf{LC}  &\textbf{mAP} &\textbf{LC} \\
      \hline
        Proto Nets  &$60.8$ &-  &$66.4$ &-  \\
        Relation Nets   &$62.1$ &-  &$67.4$ &-  \\
        LPN  &${62.3}$ &-  &${65.2}$ &-   \\
        \hline
        Proto Nets \texttt{+} NLC &${62.6}\uparrow$ &$32.2$  &$\BLUERANK{68.9}\uparrow$ &$35.1$  \\
        Relation Nets \texttt{+} NLC  &$\mathbf{64.0}\uparrow$ &$\BLUERANK{39.0}$  &$\mathbf{69.0}\uparrow$ &$\BLUERANK{35.7}$  \\
        LPN \texttt{+} NLC &$\BLUERANK{63.5}\uparrow$ &$\mathbf{42.4}$  &${66.8}\uparrow$ &$\mathbf{37.0}$  \\
        \hline
\end{tabular}
}
\caption{The accuracy (\%) of baseline methods and the additional NLC on iMaterialist. The evaluation is based on mAP and LC for 10-way 1-shot and 5-shot. The best performance is in \textbf{bold} font and the second best is in~\BLUERANK{blue} font. } 
\label{table:imatresults}
    \vspace{-0.4cm}
\end{table}

\begin{table*}[t]
    \centering
    \resizebox{0.75\textwidth}{!}{
    \small\addtolength{\tabcolsep}{-2.5pt}
    \begin{tabular}{|c|c|c|c|c|c|c|c|c|}
     \hline
     \multirow{2}{*}{\textbf{Model}} &\multicolumn{4}{c|}{\textbf{1-shot}} &\multicolumn{4}{c|}{\textbf{5-shot}}\\
    \cline{2-9}
   
    & $p1\xrightarrow{}p2$ & $p2\xrightarrow{}p3$ & $p3\xrightarrow{}p4$ & $p4\xrightarrow{}p1$ & $p1\xrightarrow{}p2$ & $p2\xrightarrow{}p3$ & $p3\xrightarrow{}p4$ & $p4\xrightarrow{}p1$ \\
     \hline
     Pre-trained \texttt{+} MLP  &$\BLUERANK{62.24}$   &$52.85$ &$66.89$  &$\BLUERANK{51.12}$ &$\BLUERANK{83.93}$ &$74.93$ &$\BLUERANK{86.93}$ &$70.63$ \\
    Proto Nets   &$58.48$   &$\BLUERANK{54.82}$ &$\BLUERANK{67.41}$  &$50.89$ &$81.52$ &$\BLUERANK{75.71}$ &$86.07$ &$\BLUERANK{72.64}$ \\
    
     Relation Nets
      &$48.50 $   &$48.74 $ &$63.03 $  &$50.93$ &$53.57$ &$66.26$   &$72.50$ &$57.20$\\
    
     {LPN}  &$\mathbf{65.41}$  &$\mathbf{60.06}$ &$\mathbf{74.60}$ &$\mathbf{53.02}$ &$\mathbf{91.34}$  &$\mathbf{79.57}$ &$\mathbf{90.89}$ &$\mathbf{77.15}$\\
    \hline
    \end{tabular}
    }
    \caption{ML-FSL results (mAP) on Open MIC for 10-way 1-shot and 5-shot. $p\{n\} \xrightarrow{} p\{m\}$ means that training is performed in {\em p\{n\}} and testing is applied in {\em p\{m\}}. The best performance is in \textbf{bold} font and the second best is in~\BLUERANK{blue} font.}
    \label{table:openmic}
    \vspace{-0.3cm}
\end{table*}

\noindent\textbf{iMaterialist.} We evaluate our ML-FSL proposal on the iMaterialist fashion
dataset~\cite{Guo2019Imaterialist} consisting more than 1 million images and 228 distinct labels. We divide  the dataset  into training, validation, and testing sets for our purpose. 
We consider the subset of the dataset of 120 labels from all labels. The splits for testing and validation are 40 and 15, respectively. The rest labels are used for training. This fashion dataset shows the  multi-label problem that has multiple labels for one object. For example, a color and a texture share the same visual appearance but they have different labels.

\noindent\textbf{Open MIC}~\cite{Koniusz2018Museum}. This dataset contains images collected  from 10 museum exhibition spaces.  There are 866 classes in total and every class comprises 1-20 images per class. The images suffer from various photometric and geometric distortions. Multi-label annotations are available as every image contains more than one exhibit. The dataset is split into four subsets: {\em p1=(shn+hon+clv)}, {\em p2=(clk+gls+scl)}, {\em p3=(sci+nat)}, {\em p4=(shx+rlc)}.

\subsection{Details}
\noindent \textbf{Implementation.} We equip all methods in our experiments  with the same variant of convolutional neural networks (CNN). The CNN has 4-convolutional layers (4-Conv) with 64 filters in each layer followed by batch normalization, ReLU, and max-pooling. %
Training and testing is performed over 100,000 and 1,000 episodes for both architectures. Adam optimizer~\cite{Kingma2014Adam}  with learning rate $0.001$ is used whose learning rate is reduced by half every 10,000 episodes. We use $\lambda=0.01$ for all methods and datasets.

We perform 10-way 1-shot and 10-way 5-shot experiments to compare the performance. The number of query images is half of the number of sampled labels. On MS-COCO, the testing stage employs a model from training for which the best validation score was attained. On the Open MIC dataset, we employ the models saved on the the last training \textit{episode}.

\noindent \textbf{Evaluation Metric.} The evaluation metric to measure accuracy in our experiments is based on mean average precision (mAP) which is the ranked list of the confidence scores compared with the list of the true labels. Another measurement is the accuracy to estimate the label count (LC).

\subsection{Results and Ablation Studies}
In our experiments, we consider the multi-label classification problem in the meta-learning setting to explore the assumption that the tasks given during training and testing are similar. The experiments are performed with baselines on the 4-Conv backbone. In addition to the baselines, we perform a comparison to the pre-trained feature extractor on the training set and a multi-layer perceptron (MLP) layer as a classifier which is updated based on the given support set. The experiments are run on 10-way and 15-way with 1-shot and 5-shot protocols as shown in Table~\ref{table:mscoco},~\ref{table:imatresults},~\ref{table:imat_15way} and~\ref{table:coco_15way}. In addition, we also provide a comparison to the existing ML-FSL method in Table~\ref{table:lasosetting}.  

\noindent\textbf{MS-COCO}. In this dataset, we evaluate the methods on the base classes and the novel classes. This protocol makes sure that the model does not only fit to the novel classes and \textit{forget} the base classes. We can observe the results from Table~\ref{table:mscoco} that pre-trained feature extractor with MLP cannot perform better than the few-shot learning baselines trained from scratch with episodic training. In all cases, label propagation network (LPN) can outperform the other methods. LPN outperforms the second highest on novel classes by 1.4\% and 1.1\% for 1-shot and 5-shot, respectively.  This is because a graph model can capture the relation among samples which share similarity.  Furthermore, there are also improvements for both base and novel classes when the basic few-shot learning methods are complemented with NLC. We conjecture that NLC imposes a regularization implicitly. On MS-COCO split by~\cite{Alfassy2019Laso}, we outperform the LASO model by 10\% and 6\% for 16-way 1-shot and 16-way 5-shot, respectively (see Table~\ref{table:lasosetting}).

\noindent\textbf{iMaterialist}. 
In this dataset, the data has a different structure compared to MS-COCO and Open MIC because two labels can share the same the same object and the labels are hierarchical. Adding a label count loss is consistently beneficial for improving the accuracy (mAP). We observe that a further improvement $\sim$1.5\% with NLC for 10-way 1-shot and 5-shot as shown in Table~\ref{table:imatresults}.  Relation Nets~\cite{Sung2018Relnet} has the highest performance for 10-way 1-shot and 5-shot. Our conjecture is that the fashion images have less tight relationship between one label to another label than images in MS-COCO. For instance, a red color can appear to any fashion entities such as dress, trouser, or shirt but it is unlikely that an apple can appear on images containing sport equipment.   

\noindent\textbf{Open MIC}. On Open MIC dataset, the evaluation is performed on 4 different exhibitions for 10-way 1-shot and 5-shot. In all cases, LPN outperforms by significant margins compared to the other three baselines with minimum of 3\% in mAP as shown in Table~\ref{table:openmic}. 

\noindent\textbf{Hard Decision}. Furthermore, we also provide the accuracy scores based on the hard decision formed by the threshold obtained from the neural label count.  Table~\ref{table:harddecision_way} shows that making hard predictions on image contents  is very challenging because we need to predict the probability of classes and decide  which final labels appear in a given image. This is non-trivial in relational learning where testing classes are disjoint from training classes.

\begin{figure}[!ht]
\vspace{-5pt}
\centering
        \includegraphics[width=.37\textwidth]{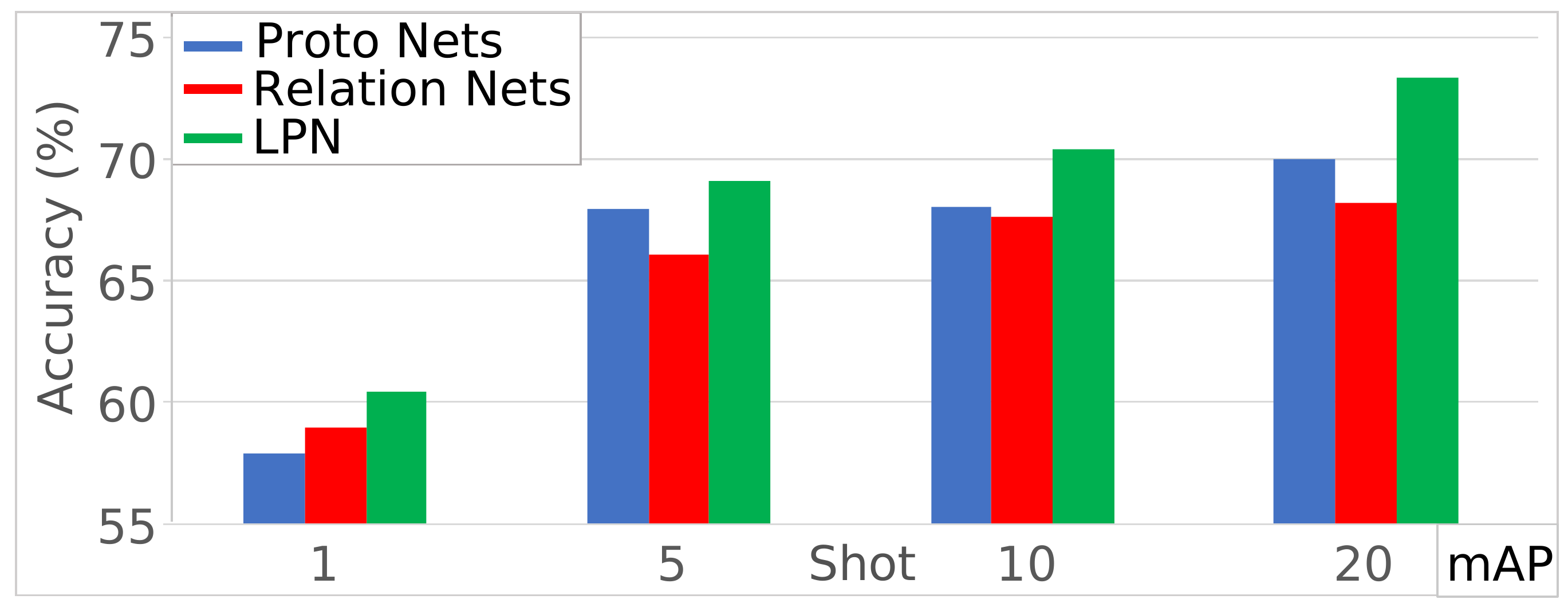}
\caption{The impact of shot on MS-COCO (mAP).}
\vspace{-0.4cm}
\label{fig:shot_map} 
\end{figure}

\begin{figure}[!ht]
\vspace{-5pt}
\centering
        \includegraphics[width=.37\textwidth]{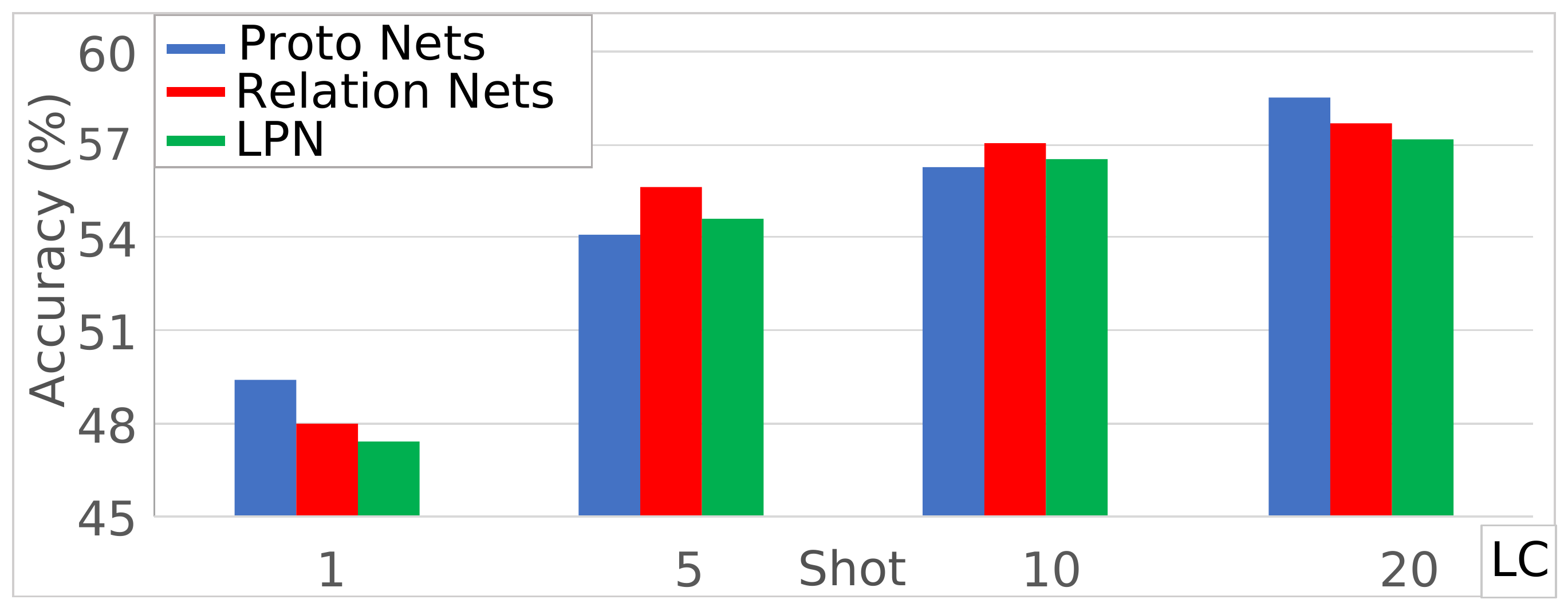}
\caption{The impact of shot on MS-COCO (LC).}
\vspace{-0.3cm}
\label{fig:shot_lc} 
\end{figure}
\noindent\textbf{The Impact of Shot.} We investigate the effect of the number of samples for mAP and LC. It is clearly shown in Fig.~\ref{fig:shot_lc} that more additional samples are beneficial to estimate the label count. The baselines are also improved when more data is given as shown in Fig.~\ref{fig:shot_map}.  We can observe that mAP and LC increase about 10\% or more from 1-shot to 20-shot.

%% file: sections/section6.tex
\section{Conclusions}
\label{sec:conclusions}

In this paper, we introduce a meta-learning framework for multi-label few-shot classification and a label counting module. The meta-learning is given in the form of \textit{episode} such that the models can gain experience by learning from past tasks and generalize to new tasks which are similar in their nature to the past tasks.  To this end, we have measured the performance on three challenging datasets: MS-COCO, iMaterialist, and Open MIC. Apart from multi-label few-shot learning protocols, we have proposed a neural label count module to estimate the number of labels. This is built based on the voting system to handle weak predictions. We showed that the module is beneficial to improve the performance even more.